\pdfoutput=1

\documentclass[11pt]{article}

\usepackage{ACL2023}

\usepackage{times}
\usepackage{CJKutf8}

\usepackage{latexsym}
\usepackage{graphicx}
\usepackage[subfigure]{graphfig}
\usepackage[T1]{fontenc}

\usepackage[utf8]{inputenc}

\usepackage{microtype}

\usepackage{inconsolata}
\usepackage{booktabs}

\usepackage{makecell}
\usepackage{multirow}
\usepackage{multicol}
\usepackage{bm}

\title{Automatic Data Visualization Generation from Chinese Natural Language Questions}

\author{Yan Ge\textsuperscript{2*}, Victor Junqiu Wei\textsuperscript{1*}, Yuanfeng Song\textsuperscript{1,3}, Jason Chen Zhang\textsuperscript{2}, Raymond Chi-Wing Wong\textsuperscript{1} \\
  \textsuperscript{1}The Hong Kong University of Science and Technology, Hong Kong SAR, China\\
  \textsuperscript{2}The Hong Kong Polytechnic University, Hong Kong SAR, China\\
  \textsuperscript{3}AI Group, WeBank Co., Ltd, Shenzhen, China \\
}

\begin{document}
\maketitle
\begingroup\def\thefootnote{*}\footnotetext{Work done during their employment at The Hong Kong Polytechnic University.}\endgroup

\begin{abstract}
Data visualization has emerged as an effective tool for getting insights from massive datasets. 
Due to the hardness of manipulating the programming languages of data visualization, automatic data visualization generation from natural languages (\emph{Text-to-Vis}) is becoming increasingly popular. 
Despite the plethora of research effort on the English Text-to-Vis, studies have yet to be conducted on data visualization generation from questions in Chinese. 
Motivated by this, we propose a Chinese Text-to-Vis dataset in the paper and demonstrate our first attempt to tackle this problem. 
Our model integrates multilingual BERT as the encoder, boosts the cross-lingual ability, and infuses the $n$-gram information into our word representation learning. 
Our experimental results show that our dataset is challenging and deserves further research. 
\end{abstract}

\section{Introduction}

Data visualization \citep{qin2020making,wang2021survey,allen2019raincloud,waskom2021seaborn} has become increasingly popular since it provides insights into data of massive size. 
In the pipeline of data visualization, an inevitable and inherent component is the creation of the \emph{specifications}, which is achieved through the \emph{declarative visualization languages} (DVL), (e.g., Vega-Lite \citep{satyanarayan2016vega} and EChart \citep{li2018echarts}). 
This DVL specifies what data is required and how the data is supposed to be visualized. 
It requires users to have expertise and knowledge of the data domain and also good programming skills of DVL, which is not quite practical, esp. for novices. 

\begin{figure}[htbp]   
	\centering

\includegraphics[width=\linewidth,scale=1.00]{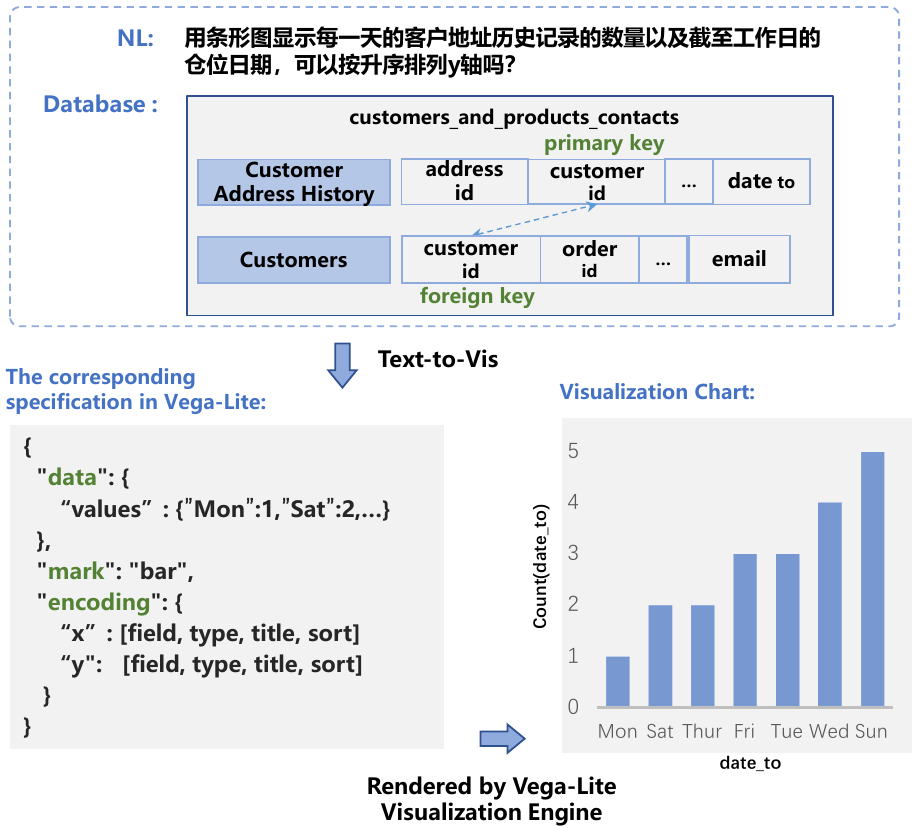}
	\caption{An example of Chinese Text-to-Vis. It should be noted that the Vega-Lite is primarily used in our discussion, due to its widespread usage and popularity \citep{song2022rgvisnet,luo2021synthesizing,qin2020making,luo2018deepeye}. However, the proposed methodology is easily adaptable to other DVLs.}
	\label{FigureOne}
\vspace{-15pt}
\end{figure}
 
Motivated by this, automatic DVL generation from natural language, or \emph{Text-to-Vis}, is becoming an emerging topic since it could provide a much more user-friendly interface. 
Many research studies have been invested in this problem, such as \citep{cui2019text,gao2015datatone,luo2020steerable,narechania2020nl4dv}. 
Given a natural language question and a database, Text-to-Vis aims to automatically translate the question into the specification in some DVLs for data visualization. 
Despite the variety of studies about this topic, we observe that all existing datasets for Text-to-Vis are for English only, and no previous studies have been conducted on Chinese Text-to-Vis datasets and methodology. 
Chinese is one of the languages that enjoy the most users worldwide. 
The lack of Chinese datasets prevents using Text-to-Vis services among these users. 
This work presents a Chinese Text-to-Vis dataset that imposes two challenges to the Text-to-Vis tasks. 
Firstly, the names of the attributes/columns in each table are typically represented in English, whereas the natural language questions are written in Chinese. 
This discrepancy requires the model to have cross-lingual ability. 
Secondly, the most basic units for denoting columns or cells can be Chinese characters, but the word segmentation can be erroneous. Figure \ref{FigureOne} is an example of the Chinese Text-to-Vis task. Given a Chinese natural language question and a corresponding database, this task aims to generate a visualization based on the semantics of the question.
This paper also presents our first attempt to tackle this problem. We adopt multilingual BERT \citep{kenton2019bert} as our encoder to boost the cross-lingual ability and infuse $n$-gram information into the word representation learning process. 
\begin{figure*}[th!]
    \centering
\includegraphics[width=1\textwidth]{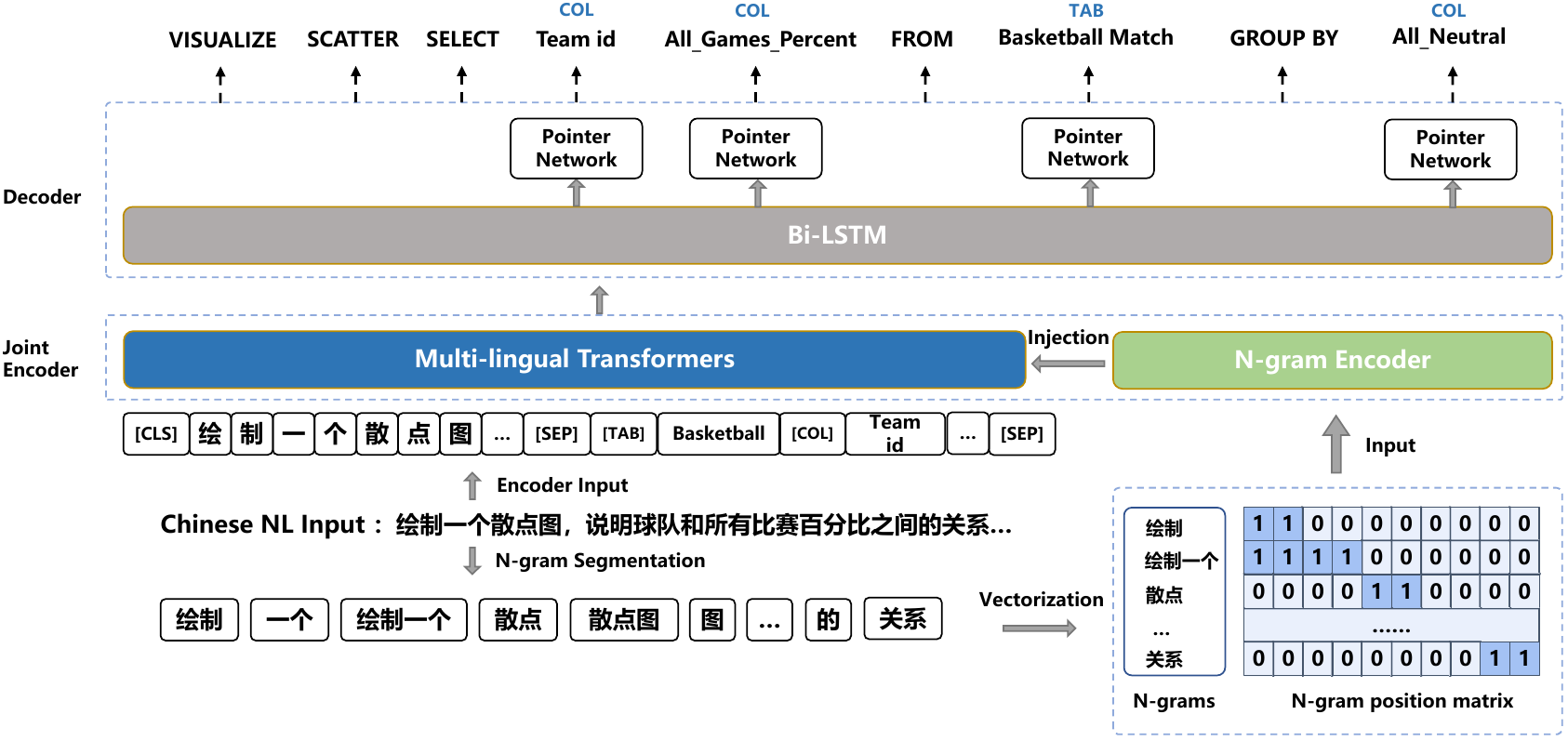}
    \caption{
   The overall structure of the proposed model, we use a cross-lingual pre-trained model to solve the language mismatch problem between natural language questions and database schema, and also integrate Chinese $n$-grams in the model, making it better able to encode Chinese semantics.}
    \label{fig:model}
\end{figure*}

In a nutshell, our contributions are summarized as follows. 
(1) We propose a Chinese Text-to-Vis dataset in this paper. To our knowledge, this is the first Chinese Text-to-Vis dataset. We detail our construction method in this paper and release our dataset to promote the development of this field. 
(2) We propose our model, the first attempt at this Chinese Text-to-Vis problem. It integrates the multilingual BERT and $n$-gram information to boost cross-lingual performance and word representation learning. 
(3) The experimental results and analysis show that our proposed Text-to-Vis task is challenging. 

\section{Related Work}

Text-to-Vis is an active area of artificial intelligence research and new techniques are emerging as the field of natural language processing(NLP) advances, it aims to convert natural language questions to visualization, making it possible for non-expert users to interact with visualization systems. There have been a variety of approaches proposed in recent years for Text-to-Vis \citep{cui2019text, moritz2018formalizing, dibia2019data2vis, luo2018deepeye, narechania2020nl4dv}, with a majority of them utilizing a learning-based approach to address the challenge. To propel the development of the data-driven solutions in this area, \citet{luo2021synthesizing} proposed a method to convert the NL2SQL dataset into a Text-to-Vis dataset and published the first large-scale Text-to-Vis benchmark NvBench. However, to the best of our knowledge, there is currently no Chinese dataset for Text-to-Vis. For a more comprehensive literature study, please refers to \ref{literaturestudy}.

\section{Dataset}

We manually translated the NvBench dataset \citep{luo2021synthesizing}   into Chinese. It should be noted that, in NvBench, both the questions and the DB (including table names, column names, and the stored values) are represented in English, but we only translated the questions into Chinese. This approach is based on the fact that professionals often construct databases using English to represent the database schema, as it adheres to programming conventions and facilitates database maintenance. In addition, the construction of this dataset aims to explore the capability of models in comprehending the semantic structure of Chinese questions and transforming them into corresponding VQL queries. This objective remains detached from the specific data languages stored within the database. The NvBench dataset includes 25,750 pairs of natural language queries and visualizations, with a total of 7,247 unique visualizations in four levels of hardness. We translate all English questions in NvBench, and we named the final Chinese dataset CNvBench.

The translation work was completed by two NLP researchers and a computer science student. The questions were first translated by one annotator, then reviewed and revised by a second annotator. Finally, a third annotator compared the original and revised versions to ensure accuracy. This process was carried out for each question to ensure the highest level of accuracy and thoroughness. When translating the questions, the translator is asked to preserve the style and structure of the original sentence if a literal translation is possible. Otherwise, If the question is complex, the translator is asked to rephrase it based on the semantic meaning of the \emph{visualization query language (VQL)} query, which is an intermediate representation of natural language question and DVL, to produce a more natural Chinese translation. We didn’t split our dataset into different subsets (e.g., training, development and test sets) since there are different perspectives of splitting the dataset, please refer to \ref{split} for details. 

\section{Method}
\label{sec:method}

In this section, we present our baseline model in response to the aforementioned cross-lingual Text-to-Vis challenges, which is inspired and inherited from the BRIDGE model \citep{lin2020bridging} due to its simple yet efficient architecture. Our model contains a BERT-based question-schema encoder for the cross-lingual encoding, following with a sequential pointer-generator to generate the corresponding VQL, which will be executed to obtain the visualization of the data. The overall structure of our model is shown in Figure \ref{fig:model}.

\subsection{Injecting N-grams information for Chinese Encoding}

 In the Chinese Text-to-Vis task, it is possible that the WordPiece\citep{kenton2019bert} segmentation (which treats each Chinese character as a token and is unaware of the boundaries of Chinese words) could cause the encoder overlooking potential database schemas mentioned in Chinese questions, preventing the model from establishing connections between them and leading to the generation of incorrect table or column names during the decoding phase. 
 
 To address or mitigate this issue, following the ZEN model \citep{diao2020zen}, we extracted $n$-grams from the Chinese question and employed an external encoder to encode these $n$-grams, then we inject the representations of the n-grams to the original cross-lingual question-schema encoder.  In detail, to encode the input  $n$-grams, a multi-layer Transformer is used as an  $n$-gram Encoder. The embedding vectors of the  $n$-grams are passed through the  $n$-gram Encoder to obtain the representation of the  $n$-grams. Then representations of each character and its associated  $n$-grams will be combined to obtain the enhanced representation, and the enhanced representation then passed to the next layer of the original encoder. This process is repeated layer-by-layer along with the original encoder. Interested readers can refer to ZEN \citep{diao2020zen} for more details.
 
\subsection{LSTM-based Pointer-Generator Decoder}
To generate the final VQL statements, we use an LSTM-based pointer-generator decoder as described in the BRIDGE\citep{lin2020bridging} model. During the generation phase, the decoder has its ability to selectively incorporate specific parts of the input sequence into the output by "pointing" to them. 
The decoder is initialized using the hidden vectors from the encoder. Then at each time step, the decoder has two options: generating a VQL keyword from the vocabulary, or using the pointer network to copy a table or column name from the schema. These options allow the decoder to create a VQL query while also incorporating relevant information from the schema.

\section{Experiments}
\subsection{Experimental setup}
We conducted quantitative experiments on both human and machine translation data to evaluate our method. In addition to the approach proposed in this paper, we also conducted experiments under a variety of settings, mainly focusing on the impact of the performance on different encoding methods in this cross-lingual task. 

In the experiment, we test the model presented in Section~\ref{sec:method}, denoted by $\textbf{BRIDGE}_{MN}$, which
integrates both the multilingual BERT and our proposed $n$-gram injection method in the encoder. 
We also test its variant $\textbf{BRIDGE}_{M}$ which only utilizes multilingual BERT but does not use the $n$-gram encoder. 

To assess the effectiveness of our proposed joint-encoder method, we also test our model with an LSTM as the encoder instead. 
But it keeps the decoder side intact. 
It adopts the Tencent multilingual embeddings\footnote{\url{https://ai.tencent.com/ailab/nlp/en/embedding.html}} as the pre-trained word embedding. 
We use two different word segmentation tools, Jieba\footnote{\url{https://github.com/fxsjy/jieba}}and HanNLP\footnote{\url{https://github.com/hankcs/HanLP}} to investigate the effect of Chinese word segmentation methods on the final results. 
We call the two models by using the two tools $\textbf{LSTM}_{J}$ and  $\textbf{LSTM}_{H}$ respectively. 

\subsection{Overall results}

  Following NvBench, we use tree matching accuracy \citep{luo2021synthesizing} and Vis matching accuracy \citep{luo2021synthesizing} to evaluate the model performance. Tree accuracy assesses the model's ability to generate the correct VQL syntax tree for a given question, and Vis accuracy reveals the model's ability to produce the appropriate visualization component. Specifically, tree matching accuracy assesses whether the VQL generated by the model is the same as the ground truth VQL, and vis accuracy focuses on the results of each component of the VQL, including the Vis type, axis and data (please refer to NvBench \citep{luo2021synthesizing} for more details). 

\begin{table}
\centering
 \setlength{\tabcolsep}{1mm}{
 \scalebox{0.83}{
\begin{tabular}{lccccc}
\hline
\textbf{} & \textbf{Easy} & \textbf{Medium} & \textbf{Hard} & \textbf{Extra Hard} & \textbf{All}\\
\hline
\bf LSTM & 0.527 & 0.534 & 0.479 & 0.486  & 0.501  \\
$\textbf{BRIDGE}_{M}$ & 0.833 & 0.824 & 0.751 & 0.760  & 0.804 \\
$\textbf{BRIDGE}_{MN}$ & 0.836 & 0.842 & 0.776 & 0.801  & 0.812  \\
\hline
\end{tabular}
}
}
\caption{\label{T1}
The overall Vis tree matching accuracy of three different encoding methods on the CNvBench. 
}
\vspace{-13pt}
\end{table}

Table \ref{T1} shows the overall Vis tree matching accuracy of our baseline model in different hardness levels. 

Our proposed model that combines Chinese n-grams performed the best and achieves 81.2\% vis tree matching accuracy overall. It also performed the best on different hardness levels. Compared to a basic multilingual BERT encoder, our $n$-gram based model achieved a nearly 1\% improvement, showing that incorporating $n$-grams into the encoder is helpful when processing Chinese. On the other hand, the model employing the LSTM as the encoder only achieved an accuracy of 50.1\%, which reflects the advantage of using current popular pre-trained language models as the encoder. Compared to LSTM, pre-trained models are expertise in modeling the context within the question and the relationship between question and schema.

\begin{table}[!htbp]
 \centering
 \setlength{\tabcolsep}{1.5mm}{
 \scalebox{0.89}{
 \begin{tabular}{lcccc}
  \hline
   \multicolumn{1}{l}{\bf{}} & \bf{Top1} & \bf{Top3} & \bf{Top5} & \bf{All} \\ \hline
   $\textbf{LSTM}_{J}$ & 0.501 & 0.537 & 0.591 & 0.681\\
   $\textbf{LSTM}_{H}$ & 0.509 & 0.529 & 0.595 & 0.652 \\
   $\textbf{BRIDGE}_{M}$ & 0.804 & 0.867 & 0.883 & 0.925 \\
   $\textbf{BRIDGE}_{MN}$ & 0.812 & 0.872 & 0.894 & 0.919  \\
  \hline
 \end{tabular}
}
 }
 \caption{\label{T2} Results on CNvBench with different model settings.}
\vspace{-6pt}
\end{table}

Table \ref{T2} summarizes the model performance in different settings. Notably, the $\textbf{BRIDGE}_{MN}$ method stands out as the most effective in capturing the semantic relationships between text and visualization. Its implementation yields the best performance with a Top1 accuracy (we use a beam search when decoding) of 0.812. Additionally, the performance of  $\textbf{BRIDGE}_{MN}$ surpasses  $\textbf{BRIDGE}_{M}$ across multiple evaluation metrics, including Top-1, Top-3, and Top-5 accuracies. This observation signifies that the N-gram injection approach enables a more comprehensive understanding of the text's underlying semantics by taking into account not only individual words but also the contextual relationships between consecutive sequences of words. 

Furthermore, since Chinese sentences need to be segmented before being processed by LSTM, we compared the effect of using two different word segmentation tools on the human-translated dataset for the questions. The results indicate different Chinese segmentation methods can affect the final results of the model due to the cumulative effect of errors while using a pre-trained model as the encoder can avoid this influence.

Please refer to \ref{Vis com} and \ref{error} for further experimental results and error analysis.

\section{Conclusion}
\label{sec:bibtex}

We construct the first large-scale Chinese sentence to Visualization dataset. We also present a strong baseline model and conduct extensive experiments in different configurations. We find that Chinese semantic parsing and cross-lingual question-schema linking are important factors affecting the experimental results. We hope that our dataset can play an active role in addressing Chinese Text-to-Vis with a data-driven approach.

\section*{Limitation}
One limitation of this study is that we only focus on the case that splitting the dataset in a question-based way. As mentioned in \citet{iacob2020neural}, there are three aspects to be considered when splitting the semantic parsing dataset,  each requiring different solutions and ideas. 

\bibliographystyle{acl_natbib}
\bibliography{custom}

\appendix
\section{Example Appendix}
\subsection{A Comprehensive Literature Study} \label{literaturestudy}
\subsubsection{Data Visualization}
Data Visualization, which converts abstract data into concrete, graphical representations, is naturally well-suited for providing an overview of large amounts of data. Data visualization can highlight patterns, trends, and relationships in the data that may not be immediately apparent from looking at raw data. To help data analysts to gain more intuitive insights from their data, the researchers in this area have done a lot of work to make it easier to convert data into visualizations. For example, Data-Driven Documents ($D^3$) \citep{bostock2011d3} is a unique approach to creating visualizations for the web that focuses on transparency and direct manipulation of the underlying data. Vega-lite \citep{satyanarayan2016vega} is a high-level language for creating interactive graphics and visualizations. It is designed to be easy to use and understand, even for users without previous experience in data visualization. VizQL \citep{hanrahan2006vizql} is a domain-specific language for data analysis and visualization. It is designed to be easy to use and understand, even for users without previous experience in data analysis or visualization. 
\subsubsection{Text-to-Vis}
 Text-to-Vis focuses on using NLP techniques to automatically generate visualizations from text data, this technique requires both natural language understanding for machine comprehension of natural language queries and translation algorithms for generating target visualizations using visualization language. DeepEye \citep{luo2018deepeye} is such a rule-based method that enables users to express their query intent using non-specific or ambiguous statements. Then the natural language input by the user is converted into an internal visualization language to generate potential visualizations.  Recently, some Text-to-Vis methods based on the state-of-the-art NLP techniques have been proposed. NcNet \citep{rocco2020ncnet} is an end-to-end solution that employs a Transformer-based model to translate natural language question to visualization. The authors proposed a novel and concise visualization grammar that enables Text-to-Vis to be performed in a machine translation way. Different from the end-to-end models, RGVisNet \citep{song2022rgvisnet} resolve the task in two phases: retrieval and revision. The authors first construct a Data Visualization (DV) codebase in advance. When a new natural language question comes, the model retrieves the codebase to find the most relevant DV query candidate as a prototype and then based on the prototype, the model revises to generate the most appropriate query.

\subsubsection{Text-to-Vis dataset}
The emergence of deep learning technology has greatly benefited the field of NLP, but the biggest obstacle currently hindering the development of deep Learning based Text-to-Vis technology is not the existence of corresponding NLP techniques, but the lack of massive data for training deep learning models. To alleviate this issue,  \citet{luo2021synthesizing}  released a public Text-to-Vis benchmark named NvBench, which contains 25,750 NL-Vis pairs across 105 domains, making it possible to use learning-based methods to solve the Text-to-Vis problem. In addition, another recent study \citep{srinivasan2021collecting} also released a curated dataset containing 893 natural language questions distributed across three datasets. However, the relatively small amount of data means that its significance is more in the field of human-computer interaction rather than constructing learning-based methods.
\begin{table*}[!htbp]
 \centering
 \setlength{\tabcolsep}{1mm}{
\footnotesize
\scalebox{0.99}{ 
 \begin{tabular}{cl|ccccccc|c|ccccc}
  \hline
  & & \multicolumn{7}{c|}{\bf Vis} & \bf Axis & \multicolumn{5}{c}{\bf Data} \\ \hline
  \multicolumn{1}{c}{}&  & \bf Bar & \bf Pie & \bf Line & \bf Scatter & \bf SB & \bf GL & \bf GS & \bf Select & \bf Where & \bf Join & \bf Group & \bf Binning & \bf Order\\
  \multicolumn{1}{c}{} & \textbf{LSTM} & 0.963 &0.954 & 0.905 & 0.921 & 0.878 &0.914 & 0.897 & 0.714 & 0.730 & 0.527 &0.651 & 0.872 & 0.652\\
  \multicolumn{1}{c}{} & $\textbf{BRIDGE}_{M}$ & 0.992 &0.974 & 0.987 & 0.959 & 0.924 &0.931 & 0.974 & 0.891 & 0.901 & 0.915 &0.870 & 0.931 & 0.893 \\
  \multicolumn{1}{c}{}  & $\textbf{BRIDGE}_{MN}$ & 0.997 &0.952 & 0.976 & 0.951  & 0.967 &0.930 & 0.955 & 0.912 & 0.895 & 0.924 &0.871 & 0.926 & 0.914 \\
  \hline
 \end{tabular}
}
}
 \caption{{\label{T3} Vis component matching accuracy on CNvBench 
}}
\end{table*}
\subsection{Dataset Split} \label{split}
To properly assess the model's performance, it is important to ensure that the data used for training is not visible to the model during evaluation. As described in NL2SQL task \citep{iacob2020neural}, we believe that there are also three aspects to be considered when splitting our dataset since both of these tasks involve retrieving data from a database. 

In the question-based split, the same VQLs are allowed to appear in different sets (e.g., training, development, or test), but the precondition is that the problems corresponding to these VQLs should not be the same. In other words, the problem statements should not overlap between the different sets, this ensures that the model is not biased towards a specific question during evaluation and can generalize to new, unseen results.  A query-based split method makes that identical VQLs do not appear in the same subset. Finally, in a database-split method, all questions related to a particular database are required to appear in different subsets. This way of splitting aims to test how well the model performs when applied to new domains, rather than just those it has seen during training. In our experiments, we only use a question-based split to evaluate the performance of our proposed baseline model.

\subsection{Results on different parts of the Vis component} \label{Vis com}
Table \ref{T3} reports the vis component matching accuracy on different encoders. Overall, the $n$-gram based encoder performs well on each vis component prediction task. When predicting the Visualization part, all three models obtained good performance, especially the bar charts. However, when predicting stacked bars, the results of all three models decreased compared to other Vis parts. This is because stacked bars are sometimes mentioned implicitly in the question, requiring the model’s ability to parse the mentions based on the sentence context.  For predicting the axis parts, only the LSTM encoder model obtains a poor result, the reason is that there are some corresponding aggregate functions occurring in the Select parts, and the LSTM encoder is not able to well capture this type of information in the question. For data parts, both models based on pre-trained encoders achieved good results, with LSTM still performing the worst in this part.

\subsection{Error analysis} \label{error}
To identify the causes of errors, we conducted an error analysis on our test set of 2562 VQL examples. Utilizing $\textbf{BRIDGE}_{M}$,  we identified several sources of errors from the 501 failed examples out of 2562. 

For about 39 examples, the model produces wrong predictions for Vis part. For example, the model produced a wrong VQL for the question "\begin{CJK}{UTF8}{gbsn}
所有3到5颗星的电影的片名有多少？并按降序显示名称。
\end{CJK}",  the model incorrectly predicted the visualization type as "pie" when it is actually "bar", this is due to the lack of explicit mention of the visualization type in the question. Additionally, due to the uneven distribution of visualization types among the total train samples, the model may perform well on the majority types but poorly on others.

For about 305 examples, the model generates wrong column names or table names in axis part. For example, considering the question "\begin{CJK}{UTF8}{gbsn}
关于日志中记录的不同故障描述的故障数量，按柱状图从低到高的顺序排列。
\end{CJK}", the model made a wrong prediction on column name "fault\_description" as "fault\_status", in addition to errors in predicting the column or table name, the model may also make wrong predictions on the number of column names or table names and insert extra ones into the VQL.

Errors in the data part of the VQL mean that the model makes mistakes in predicting the keywords "where", "group", "bin", and "order" of the VQL. There were a total of 224 samples with errors in this part. For the quesiotn 
"\begin{CJK}{UTF8}{gbsn}
对于名字中有字母D或S的所有员工，请给我比较一下经理id在雇佣日期和工作日的总和。
\end{CJK}", the model made the mistake of predicting "bin by weekday" as "bin by month". 
\end{document}